# A Model-Based, Decision-Theoretic Perspective on Automated Cyber Response

**Lashon B. Booker    Scott A. Musman**

The MITRE Corporation, 7515 Colshire Drive, McLean VA 22102

booker@mitre.org  ,   smusman@mitre.org

## Abstract

Cyber-attacks can occur at machine speeds that are far too fast for human-in-the-loop (or sometimes on-the-loop) decision making to be a viable option. Although human inputs are still important, a defensive Artificial Intelligence (AI) system must have considerable autonomy in these circumstances. When the AI system is model-based, its behavior responses can be aligned with risk-aware cost/benefit tradeoffs that are defined by user-supplied preferences that capture the key aspects of how human operators understand the system, the adversary and the mission. This paper describes an approach to automated cyber response that is designed along these lines. We combine a simulation of the system to be defended with an anytime online planner to solve cyber defense problems characterized as partially observable Markov decision problems (POMDPs).

## Introduction

Cyber analysts are faced with a daunting set of challenges as they try to craft responses to increasingly sophisticated cyber-attacks. Typically, analysts are overloaded with too many diverse and noisy alerts to process, making it difficult for them to adequately assess the cyber situation. This means they often must rely on incomplete and uncertain information as a starting point for making decisions about how to act. It also means that analysts can struggle to find coherent response sequences that address the broad spectrum of alerts received. In order to trace suspicious events to a root cause, it is often necessary to correlate information across multiple event streams and over multiple temporal windows. Moreover, analysts often don't understand the implication of their actions in terms of mission success or failure for the system being defended. This is all complicated by the fact that a timely response can be problematic when attacks occur at machine speeds. Much of what occurs today relies on pre-determined responses to contingencies, seat-of-the-pants decisions, and sometimes knee-jerk reactions that may result in response actions that are worse than the attack itself.

Many applications of AI to cyber security problems are focused on helping analysts manage these challenges. There is a case to be made, though, that even with AI support, current approaches to cyber security might be overwhelmed by a new generation of AI-enabled attacks. Future cyber-attackers are likely to increasingly exploit advances in AI to achieve faster, stealthier, and more effective operational effects. Many of these effects will be achieved at a speed and scale that makes a human-in-the-loop defense paradigm unlikely to be effective. Consequently, future systems will have to rely to some extent on automated reasoning and automated responses – with humans on the loop or out of the loop – to ensure mission success and continuously adapt to an evolving adversary.

If AI takes on more responsibility for reasoning and response in cyber defense, what are the implications for human-machine teaming? One clear implication is that AI cannot be viewed as just a tool applied by human operators to make cyber problems more tractable. AI becomes an agent that acts with human partners and on their partners behalf. If this partnership is going to succeed, humans must be comfortable that the AI agent will help achieve their goals in situations where the system mission is at risk, even under uncertainty. The underlying issues here are dependability and building trust.

Three attributes of an automated agent are particularly important regarding these issues (Lee & See, 2004): purpose, process, and performance. Purpose refers to the designer's intentions in building the system and the degree to which the system is being used in accordance with those intentions. Process refers to an understanding of the factors (rules, control laws, algorithms, etc.) that govern system behavior. Performance refers to the demonstrated ability to achieve desirable goals (competence, reliability, predictability, etc.).

This paper describes an approach to automating cyber response that is designed with these attributes in mind. We start with the premise that, from an AI perspective, it is advantageous to frame the cyber response problem as a sequential decision-making problem under certainty. This leads naturally to using decision-theoretic approaches to represent the way a human operator understands the system, the adversary, and the mission; and generate responses that are aligned with risk-aware cost/benefit tradeoffs defined by user-supplied preferences. The result is an automated AI agent that is well-suited to fill the role of a dependable and trustworthy partner for human operators.

## Managing Uncertainty in Cyber Defense

As we have defined it, automated reasoning about cyber responses is essentially decision-making based on the projection of possible futures from a current situation. Viewed as a form of game-playing, this involves sequential decision making where the defender and attacker are each afforded an opportunity to make a move or do nothing (a "no operations" or NOP). The complication in cyberspace, however, is that information about the current system state and future projections or attacker actions are highly uncertain.

One way to account for these issues is to address the cyber response problem directly as a partially observable stochastic game (e.g. as a partially observable competitive Markov decision process (Zonouz, et al., 2014)). However, suitable state-of-the-art solution techniques are only capable of solving relatively small games that must be fully specified in advance. Moreover, while optimal solutions – when tractable – compute the Nash equilibrium that specifies optimal policies for both the attacker and defender, these policies are conservative in the sense that they do not necessarily exploit opponent weaknesses. It is not clear that this kind of conservative strategy is always the most effective approach to responding to a cyber attack.

One alternative to a game-theoretic solution is to focus on resolving the defender's uncertainty about how to respond, rather than trying to solve the complete stochastic game. When the opponent's policy is fixed (either known or estimated from data), we can model a partially observable stochastic game as a partially observable Markov decision problem (POMDP) from the perspective of the protagonist (Oliehoek, et al., 2005). The adversarial aspects of the stochastic game are incorporated into the transition function of the POMDP. This is an attractive option because recent advances in POMDP solution techniques make it possible to solve large-scale POMDPs in real time. Additionally, POMDP solvers can find policies that exploit opponent weaknesses. For these reasons, our research tackles the cyber response challenges using the formal framework of partially observable Markov decision problems. Note that the POMDP approach can compute the kind of general-purpose conservative solution one would expect from a game-theoretic approach if we formulate the POMDP to assume a robust adversary like a min-max opponent.

## Partially observable Markov decision problems

Formally, a POMDP can be expressed as a tuple (S, A, Z, T, O, R) where S is a set of states, A is a set of actions, Z is a set of observations, T(s, a, s') is a transition function giving the probability p(s' | s, a) of transitioning to state s' when the agent takes action a in state s, O(s, a, z) is an observation function giving the probability p(z | s, a) of observing z if the agent takes action a and ends in state s, and R(s, a) is a reward function giving the immediate reward for taking action a in state s. The goal of the decision maker is to maximize the expected reward accrued over a sequence of actions. A solution to the decision problem is an optimal policy that specifies a mapping from states to actions which can be used to guide action choices and achieve the maximum expected total reward. Since the states in a POMDP are not fully observable, the only basis for decision making is the sequence of prior actions and subsequent observations. A sufficient statistic summarizing the probability of being in a particular state, given a history of actions and observations, is called a belief, and a probability distribution over all states is called a belief state. Solving a POMDP is a planning problem that involves finding a policy which maps belief states to actions.

Clearly, any search involving probabilistic belief states and arbitrarily long histories of actions and observations quickly becomes computationally intractable (Pineau, et al., 2003) because of the "curse of dimensionality" and the "curse of history". POMDP algorithms typically mitigate these concerns by avoiding exact representations of the belief space and working instead with samples from the belief space to compute good approximate solutions. For example, many algorithms compute solutions by estimating the optimal value function associated with the optimal policy. The value function for a policy provides an estimate of the reward expected from executing the policy from a given belief state. Value iteration techniques have long been relied on to efficiently compute these value functions for POMDPs (Cassandra, et al., 1994) and they provide the basis for a variety of offline solution methods that compute a value function over the entire belief space before taking an action. The best current exemplar of this approach is probably the SARSOP (Kurniawati, et al., 2008) algorithm, which can find good solutions to POMDPs having up to 10,000 states in a practical amount of time.

Although state-of-the-art offline methods for solving POMDPs have made great strides, they are not yet powerful enough to address the challenges of real-world cyber response problems. Fortunately, there are approaches available to (sometimes approximately) solve POMDPs online in real time that appear to be suitable for our purposes.

**Online planning for solving POMDPs**

An alternative to offline planning is to select actions online, one at a time, using a fixed-horizon forward search (Ross, et al., 2008) (He, et al., 2011). Here, the key to making this idea effective for real-world problems relies on sampling the belief space, rather than fully exploring it. In particular, great efficiencies can be achieved by using a black-box simulator of the POMDP in conjunction with sampling. The partially observable Monte-Carlo planning (POMCP) (Silver & Veness, 2010) algorithm uses a Monte Carlo tree search with weighted rollouts to sample belief states and state transitions in the belief tree (thereby mitigating the curse of dimensionality), along with a black box simulator of the POMDP to estimate the potential for long-term reward (thereby mitigating the curse of history). Belief states are efficiently approximated and updated using particle filters. POMCP was the first general purpose planner to show how POMDPs with state spaces as large as $10^{56}$ could be solved with only a few seconds of computation. The algorithm provably converges to the optimal value function of the POMDP if the beliefs updated from the Monte Carlo samples accurately reflect the action/observation history.

While the POMCP algorithm can achieve impressive results on some large POMDPs, the extremely poor worst-case behavior of its lookahead search makes it a questionable choice for cyber security problems. The DESPOT algorithm (Ye, et al., 2017) is an anytime online algorithm for POMDP planning that avoids the worst-case behavior of POMCP. Rather than using weighted Monte Carlo rollouts to sample a belief tree, the idea is to generate a randomized but systematically extracted sparse subtree of the belief tree. This sparsely sampled belief tree is called a DESPOT (Determinized Sparse Partially Observable Tree). A DESPOT captures the performance of policies over a random but limited sample of "scenarios" by including all possible action branches in the corresponding belief tree but only including those observation branches that occur in a sampled scenario. The DESPOT is constructed incrementally using heuristic search with branch and bound pruning, expanding the tree selectively in the most heuristically promising direction on each step. Theoretical results show that, given a suitable number of scenarios to work with, the DESPOT algorithm can reliably find near optimal policies with a regret bound that depends on the size of the optimal policy. This approach has been successfully applied to compute real-time solutions to complex POMDP planning problems for autonomous vehicles. Its performance characteristics, and its characteristics as a decision-theoretic planner (Boutilier, et al., 1999), make it a good choice as the starting point for building a POMDP planner to address cyber security problems.

**A simple cyber defense scenario and POMDP**

To allow us to systematically explore solutions to the cyber security problems we describe, we have formulated the following notional scenario.

In Figure 1, we show a small micro network containing an attacker start point, and two target nodes (t1, t2) that can be compromised to cause mission impact, each with a middle node (m1, m2) separating the attacker from the target. While operating the network, the defender obtains intrinsic rewards for every time step that doesn't involve a compromised target host. When an attacker compromises a target host, it causes an adverse mission impact for the defender. The defender is provided a set of sensors that make it possible (at times imperfectly) to assess system state and detect attacker actions. Additionally, the defender is provided a set of response actions which can be used to restore aspects of the system to an uncompromised state. Each of these response actions have an associated cost to employ.

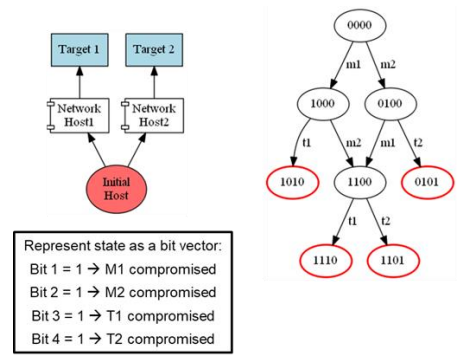

Figure 1 - Simplified micro-network

With this definition of the problem as a starting point, we construct a POMDP specification for the defender's decisions regarding an instance of our micro-network as follows. There are 4 nodes in the network, and each node can either be compromised or uncompromised. Thus, the network state can be represented as a 4-bit binary string with a "1" in a specific bit designating that the corresponding node is compromised, and a "0" in that bit designating that the node is uncompromised. Any state containing a target that has been compromised is considered a terminal state.

There is a single IDS sensor reporting the state associated with each node, and we assume they report the binary status of a node as "good" or "bad". These sensors operate independently, are reset after each attacker move, and are characterized by a false negative rate and a false positive rate.

Observations for the defender are 4-bit binary strings showing the (possibly erroneous) sensor returns.

In this scenario, there are four defender actions, consisting of two targeted actions (*Rm1* - reset m1; *Rm2* - reset m2), one global action (*RA* - reset all hosts), and the option to do nothing (*NOP*). These actions are somewhat deliberately selected such that, *Rm(1 or 2)* represents a targeted surgical response that works well when sensors can identify the hosts that have been compromised, while RA is a useful failsafe response that can reset the whole system to a safe state even when we have sensors that are unable to detect when hosts are compromised. Note, as is analogous to reconstituting a real computer to a known good state, every action which resets a node also blocks access to that node for the attacker during that stage of the game. This means, for example, that if the attacker has compromised m1 and is preparing to use it as a starting point to attack t1, as long as actions *Rm1* or *RA* are started before the attack can complete, they will foil the pending attack and the attacker must start over again from the beginning.

The reward function is given by a set of utilities which were chosen, somewhat arbitrarily, to characterize potential tradeoffs between the defender actions for this network. The defender gains 10 points for every game step in a non-terminal state, loses 800 points for entering a terminal state, loses 30 points for performing actions *Rm1* or *Rm2*, and loses 50 points for performing action *RA*. We assume the attacker scores the game the same way the defender does (i.e., a zero-sum game) and the cost of performing attacker actions does not currently factor into the total score.

Despite its simplicity, this micro network provides a useful starting point for assessing automated response solutions since it incorporates multi-stage attacks, probabilistic actions, and uncertain sensing. One example of the kind of investigation that is instructive involves sensor noise.

The optimal policies for some scenarios involving small amounts of sensor noise can be surprisingly complex. In the case of an attacker who uses a greedy strategy, this becomes especially prevalent when different kinds of sensor noise are mixed together, or when sensor noise is combined with uncertainty about when the attacker will make a move. The subtleties of such complex decisions are likely beyond the capabilities of a human decision maker who is forced to make a hurried decision.

Figure 2 shows the policy graph for an example involving a relatively small (0.1) false positive rate together with a somewhat large probability (0.9) that the attacker will choose the NIL action and hence the defender can wait to make an overt response. A policy graph (Cassandra, et al., 1994) summarizes the action choices made by the optimal policy. Each node corresponds to a distinct belief state and shows the system state having the greatest belief, along with its belief value, and the associated optimal action. Outgoing edges indicate expected observations, along with their probability. The policy graph in Figure 2 shows how the optimal[1] policy for this problem instance resolves the uncertainty about several contingencies. Note, in particular, the long chain of nodes in the middle that address the possibility that repeated false positive errors could hide the presence of a compromised asset. This is one of the more straightforward examples of the complexities we have found in solutions to our simplified cyber security problem (Musman, et al., 2019). Some solutions involve policy graphs having over 100,000 nodes and more than 300,000 edges.

## Modeling the Cyber Terrain and Attacker Behavior in CSG

In our approach, the black box simulator needed in conjunction with the online planning paradigm is provided by the Cyber Security Game (CSG) (Musman & Turner, 2018). CSG provides a coarse-grained simulation of attacker and defender interactions in cyberspace.

The original implementation of CSG focused on assessing defensive architectures and the deployment of static cyber defenses. CSG's cyber mission impact assessment (CMIA) model (Musman, et al., 2010) (Musman & Temin, 2015) quantifies the consequences (expected loss) in the risk equation[2]. The probability of compromise used in the risk equation is approximated by computing the difficulty of traversing pathways through the cyber terrain from attacker footholds. Defensive decisions are assessed by running a simulated attacker against defender option combinations to determine if the risk scores have been reduced. In the original version of CSG, the attacker is assumed to be omnipotent, having a full knowledge of the target network. Hence for each CSG move pair, the computed risk score is based on the attacker following a min-max strategy to cause the worst impacts, using the best attack pathways to reach the assets that cause impacts.

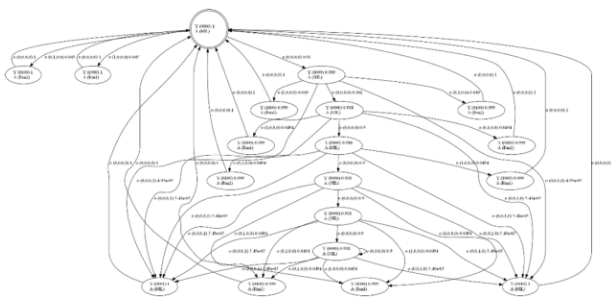

Figure 2 - Policy graph for a problem instance with false positive errors and long attacker dwell times

---

[1] We computed the optimal policy using the SARSOP algorithm (Kurniawati, et al., 2008).
[2] CSG defines individual incident risk as the product of the probability that a cyber incident will occur (i.e., compromises) and the expected loss incurred from the incident (i.e., consequences). CSG then defines the total system risk as the summation of all the incident risks associated with the possible set of incidents that an attacker can cause.

CSG's defensive cyber decision making focuses primarily on defending the mission that the cyber assets are intended to support. This mission focus helps reduce the scope of the cyber defender's problem since often only a subset of the cyber assets is relevant at any given time. CSG uses a CMIA model to translate the occurrence of incidents in cyberspace into mission outcome impacts (whether they be functional loss, financial loss, loss of reputation, loss of life, etc.). In a CMIA model, cyber assets are treated like any other organizational resource that exists to support mission related activities. Activities that depend on cyber assets are typically assumed to be performed correctly unless the cyber assets have been affected by some cyber compromise.

A challenge in cybersecurity is to be able to comprehensively consider the potential impacts of what is a staggering number of exploits and cyberattack methods (e.g., CVE has over 80,000 entries and CAPEC enumerates over 500 cyber attack patterns (MITRE Corporation, 2013)). To avoid having to reason about every possible attack instance, CSG's approach is to reason about the effects of successful attacks, rather than the attack instances themselves.

The effects of cyber compromises are represented by the set of incident effects in the DIMFUI (Temin & Musman, 2010) taxonomy. These effects are defined in Table 1. The DIMFUI effects were chosen to provide a robust representation of cyber incidents. Every successful cyber compromise that exists in CVE, and which is described by a CAPEC attack pattern, can be represented by one or more DIMFUI effects against one or more cyber assets in a system. More recently, the MITRE ATT&CK (MITRE Corporation, 2019) framework has provided a more operational mapping of the techniques used by malicious cyber actors. From a DIMFUI perspective, many of the techniques listed in ATT&CK belong to the category of unauthorized use in terms of asset compromise, lateral movement within a compromised network, and privilege escalation. Of the DIMFUI incident effects, degradation is the only one that involves a range of values and that range is something that must be understood specifically in terms of the type of cyber asset (e.g., communications channel, CPU, data) and the mission it supports. All of the other DIMFUI effects represent the binary states of an ICT asset. A cyber adversary is either able to modify a cyber asset or they are not. An ICT asset is operationally available, or it is not; a component or piece of data has been fabricated, or it has not, etc. This makes DIMFUI a useful abstraction that allows a cyber defender to reason only about the impact of 6 DIMFUI incident effects, rather than 100's or thousands of attack instances.

From a security monitoring perspective, because the CMIA model captures which types of incident effects cause mission impacts, there is an opportunity to make security instrumentation more precise and targeted. For example, if it is known that information must be available, the availability status of the information asset can be monitored. If it is known that information has integrity requirements, then substantiated integrity techniques enable the integrity of the asset to be monitored. This provides significant context to understand which security events are relevant and related to impact situations.

CSG relies on models of the cyber terrain, mission impact, attacker, and defender capabilities. A typical cyber model used in CSG (shown in Figure 3) consists of networks, network components (i.e. switches, routers, firewalls), hosts on the networks, user groups having access to the hosts, peripherals, applications, services and interactors that run on the host, and information used in the performing mission function. We refer to the system topology model as a cyber terrain model, since it contains trust and access relationships in addition to physical connectivity. The existence of user groups, that may have access to multiple assets in the network, provides a way to simulate how compromised user credentials can be used to access hosts.

Table 1 - The DIMFUI taxonomy

| DIMFUI | Explanation | Typical Attacks |
| --- | --- | --- |
| Degradation | 1. Reduction in performance or capacity of an IT system<br>2. Reduction in bandwidth of a communication medium<br>3. Reduction in data quality | 4. Limited-effect DoS<br>5. Zombie processes using up CPU and slowing server<br>6. Transfer of non-mission related data over a link that slows the transfer of mission data<br>7. Dropped packets cause an image to have less resolution |
| Interruption | IT asset becomes unusable or unavailable | 1. Ping of Death<br>2. Wireless Jamming<br>3. Wipe disk |
| Modification | Modify data, protocol, software, firmware, component | 1. Change or corrupt data<br>2. Modify access controls<br>3. Modify/Replace system files |
| Fabrication | Attacker inserts information into a system or fakes components | 1. Replay attacks<br>2. DB data additions<br>3. Counterfeit software/ components |
| Unauthorized Use | Attacker uses system resources for illegitimate purposes. Related and often a precondition for other DIMFUI. | 1. Access account or raise privileges in order to modify/degrade/interrupt the OS<br>2. Subvert service to spawn a program on remote machine<br>3. Bandwidth used surfing for porn degrades mission critical exchanges |
| Interception | Attacker gains access to information or assets used in the system | 1. Keylogger<br>2. SQL injection<br>3. Crypto key theft<br>4. Man-in-middle attacks<br>5. Knowledge of component or process that is meant to be secret |

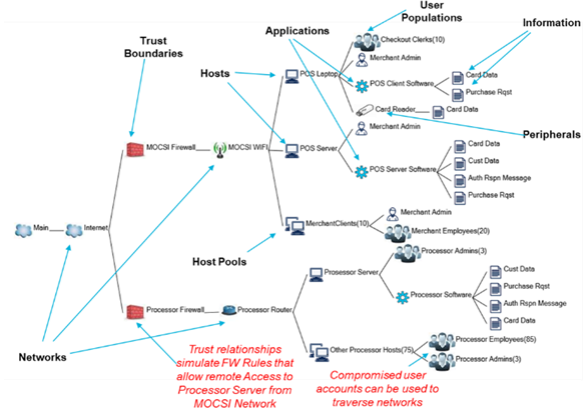

Figure 3 - Typical level of detail in CSG cyber models

The original implementation of CSG was used to represent a fully-observable, probabilistic outcome, zero-sum game for assessing the employment of static defenses. For the purposes of supporting our work on online planning, CSG has been modified to support the ability to interleave attacker/defender moves within a single game simulation, playing forward to some outcome where either the attacker succeeds in creating a mission impact or has been fended off by the defender.

## Online Planning for Automated Cyber Defense

Our previous analysis of the optimal POMDP solutions for even simplified cyber security problems showed how quickly the decisions the defender must make become too complex for humans to easily develop on their own. This underscores the need to automate these decisions and find good approximate solutions to large-scale POMDPs in real time. Our work on Automated Reasoning about Cyber Response (ARCR) is a step toward addressing that need by combining the capabilities of CSG with an online planner.

We used the Approximate POMDP Planning (APPL) toolkit[3] to build an online planner that employs the DESPOT algorithm. This toolkit makes it possible to implement a customized planner that includes problem-specific heuristic bounds on forward search, arbitrary representations for POMDP states, beliefs and observations, and a clearly defined interface for our black-box simulator.

The current implementation of our ARCR planner uses custom bounds to manage policy search derived from knowledge about the cyber problem at hand. In the micro network scenario, for example, we know that performance will be bounded from above by policies using the surgical responses of *Rm1* and *Rm2* when we have perfect sensors, and bounded from below by policies using the *RA* response when we have useless sensors. Note that one of the advantages of having CSG available as our simulator is that we can run it offline to estimate heuristic bounds for the planner when the problem is too complex to derive the bounds we need directly. Because most real-world computing networks are relatively flat, deep lookaheads are not needed to generate these estimates. Cybersecurity attack trees tend to be broad, rather than deep, so a response policy heuristic that favors simple short response strategies over complex long ones is likely to be effective, even if the strategies are not always optimal. Our future work will explore how well these hypothesized heuristic bounds work in practice.

[3] http://bigbird.comp.nus.edu.sg/pmwiki/farm/appl/

Because cyber problems involve events at multiple time scales, the ARCR planner uses macro-actions (He, et al., 2011) to make long multi-step lookahead searches more efficient. We have implemented what we call "discrete event" macro-actions which replace sequences of primitive steps where the attacker is idle with a single lookahead step whose expected reward is calculated analytically. These macro-actions provide a significant speed-up in planner performance.

Sensor noise can make the game tree searches in CSG intractable by greatly increasing the branching factor, just as it can make the forward search of belief states intractable in the planner. Since the planner already includes heuristics and bounds to manage the complexity of the search, we decided to implement the sensor models in the planner. Our sensor model implementation exploits the assumption that sensor error events are independent to compute the observations required on each planning step efficiently.

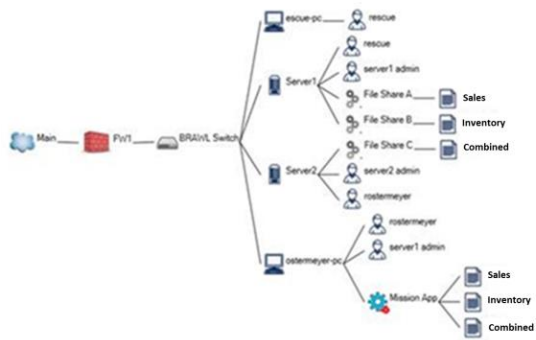

Figure 4 - A simple use case illustrating an information fusion mission

Our current work is applying the ARCR planner to more realistic cyber defense problems that involve several DIM-FUI effects. Figure 4 shows a simple use case involving an information fusion mission. Business transaction agents (not shown) generate Sales and Inventory files that are placed in File Shares A and B respectively being served from Server

1. A client agent accesses paired Sales and Inventory files, performs some (unspecified) fusion operation on them and produces a combined status update file as an output, which is placed in Shared Folder C being served on Server 2. It is presumed that there is mission value to generating the combined status files in a timely fashion, while maintaining their integrity and confidentiality.

In our initial experiments with this use case, we modeled a persistent attacker with a greedy strategy who causes impact as soon as possible. The attacker steals a user credential on its foothold, then uses that credential to move laterally from the foothold to Server 1. Once on Server 1, the attacker modifies the file share and modifies one (or more) files on the file share, thereby causing adverse impact to the mission.

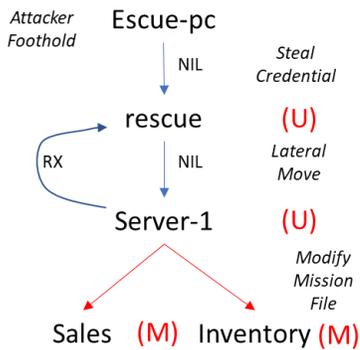

Figure 5 - Interaction between attacker and defender when the lateral move is detected

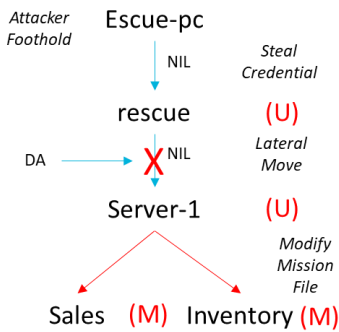

Figure 6 - Interaction between attacker and defender when defender can disable accounts

Figure 5 illustrates the interaction between the attacker's actions, the associated DIMFUI effects, and the defensive responses generated by the ARCR planner. Assuming the available sensors do not detect the stolen credentials but do detect the lateral move, the easiest response is to eject the attacker and prevent impact by restoring the host with RX. While this response defends against the attack, it does not eliminate the threat and the attacker can simply go after the host again.

If the defender is provided with an action that can disable a user account (DA), the planner can determine that the DA action completely blocks the attacker from doing any damage and is therefore the preferred solution (assuming that disabling accounts is not too costly). This defensive response is illustrated in Figure 6.

We are currently investigating how the defensive policy changes given different assumptions about what the sensors can detect, how reliable the sensors are, and how variations in vulnerabilities and credential use affect the attack paths.

## Summary

Future cyber-attackers are likely to increasingly exploit advances in AI to achieve faster, stealthier, and more effective operational effects. Many of these effects will be achieved at a speed and scale that makes a human-in-the-loop defense paradigm unlikely to be effective. Consequently, future systems will have to rely to some extent on automated reasoning and automated responses – with humans on the loop or out of the loop – to ensure mission success and continuously adapt to an evolving adversary.

This paper describes research suggesting that it is feasible to address this challenge by using decision-theoretic techniques to build an automated, rational AI agent that can work with human analysts to achieve shared goals in uncertain situations where the system mission is at risk. Decision-theoretic approaches can represent the way a human operator understands the system, the adversary, and the mission; and generate responses that are aligned with risk-aware cost/benefit tradeoffs defined by user-supplied preferences. The result is an automated AI agent that is well-suited to fill the role of a dependable and trustworthy partner for human operators.

Our work on Automated Reasoning about Cyber Response (ARCR) has taken several successful steps in this direction. By framing the cyber response problem as a POMDP, we bring together state-of-the-art techniques for anytime online planning in large state spaces with the capabilities for modeling cyber security problems found in the Cyber Security Game (CSG). This combination appears to be a promising path toward computing tractable solutions to complex cyber security problems. Human operators could delegate responsibility for some aspects of cyber defense completely to automated responses computed in this way; or, they could use automated responses to buy time and limit loss/damage while analysts assess the situation and consider their options.

As noted previously, three attributes of an automated agent are particularly important regarding the issues of dependability and building trust with human operators: purpose, process, and performance. The decision-theoretic underpinnings in the ARCR planner and CSG provide a useful starting point for establishing strong capabilities addressing each of these attributes.

CSG was designed to assess mission risk in cyber security systems. It's explicit representation of the mission goals, and the use of DIMFUI events as an abstraction for characterizing mission-relevant outcomes, provide a useful way for humans to express their intentions. The same representations also allow us to align ARCR's behavior with those intentions to establish a shared sense of purpose.

Several factors that control the behavior of CSG and the ARCR planner are exposed to help make the decision-making process more transparent to the user. In addition to providing their mission-related preferences and utilities, humans can specify prior beliefs about system state, customized macro-actions indicating preferred/required policies or standard operating procedures for various conditions, and heuristic guidance constraining the solutions to be considered.

Finally, the performance of the ARCR system – in terms of competence, reliability, and predictability – is well characterized because actions are derived from rational decisions made in accordance with decision-theoretic principles.

Our future work will focus on expanding these capabilities and making them more efficient. Ongoing work includes the addition of state abstraction hierarchies to improve the scaling properties of the POMDP solution approximations. Other ongoing work will increase the realism of our cyber models by adding more sensors (including the detection of both action and state), adding more attacker and defender actions, and reasoning about the impacts from multiple incident effects.